\documentclass[runningheads]{llncs}

\usepackage[T1]{fontenc}
\usepackage{amsmath,amssymb}
\usepackage{mathtools}
\usepackage{bm}
\usepackage{graphicx}
\usepackage{booktabs}
\usepackage{longtable}
\usepackage{multirow}
\usepackage{subcaption}
\usepackage[hidelinks]{hyperref}
\usepackage{xcolor}

\newcommand{\R}{\mathbb{R}}

\newcommand{\KL}{\mathrm{KL}}
\newcommand{\vt}{\bm{v}_{\bm\theta}}

\newcommand{\latent}{\bm{z}}
\newcommand{\cond}{\bm{c}}
\newcommand{\policy}{\pi_{\bm\theta}}
\newcommand{\refpolicy}{\pi_{\mathrm{ref}}}

\begin{document}

\title{CONFLUX: A Latent Diffusion Model for \\
3D Chest-CT Synthesis with RL Post-Training}
\titlerunning{Latent Diffusion for 3D Chest-CT Synthesis with RL Post-Training}

\author{Max Van Puyvelde\inst{1,2}\thanks{M.~Van Puyvelde and H.~I.~Gulluk are joint
first authors; W.~Van Criekinge and O.~Gevaert are joint senior authors.} \and
H.~Ibrahim Gulluk\inst{3} \and
Wim Van Criekinge\inst{2} \and
Olivier Gevaert\inst{1}}
\authorrunning{M.~Van Puyvelde et al.}
\institute{Department of Biomedical Data Science, Stanford University School of
Medicine \and
Department of Mathematical Modelling, Statistics \& Bioinformatics, Ghent
University \and
Department of Electrical Engineering, Stanford University\\
\email{maxvpuyv@stanford.edu, gulluk@stanford.edu,
wim.vancriekinge@ugent.be, ogevaert@stanford.edu}}

\maketitle

\begin{abstract}
Controllable generative models of 3D medical images can synthesize volumes with
specified clinical attributes, but this demands samples that are simultaneously
high-fidelity, natively 3D, and faithful to the requested conditioning. We present
CONFLUX, a latent diffusion model for chest computed tomography (CT): a 3D
variational autoencoder compresses each volume, and a rectified-flow transformer
generates in the latent space. Generation is conditioned on structured radiological
metadata ($18$ abnormality findings, sex, age, and reconstruction kernel) through
adaptive layer normalization. The model leads strong volumetric baselines on
tri-planar Fr\'echet distance (FID $32.3$ vs.\ $74.6$ for MAISI) while exposing
direct control over clinical attributes. To
strengthen that control we add an online reinforcement-learning post-training stage
(group-relative policy optimization) that rewards how reliably a classifier
recovers the requested findings from each generated volume. Judged by a separate,
independent classifier, post-training removes $47\%$ of the shortfall relative to
real-scan reliability. We release the model and a ${\sim}200$k synthetic chest-CT
dataset with conditioning metadata spanning a wide variety of clinical findings.
\end{abstract}

\section{Introduction}
\label{sec:intro}

Generative models of 3D medical images are used to augment under-represented
cohorts, share data under privacy constraints, and synthesize counterfactual
volumes. Each use requires the model to be controllable: a sample requested
to exhibit a set of clinical attributes must realize them. Chest CT, which is volumetric,
high-resolution, and described by structured radiological metadata, is an especially
hard case; a model that is both high-fidelity and controllable on it is directly
useful for cohort augmentation and controlled study design.

High-resolution synthesis has converged on a latent rectified-flow design: an
autoencoder compresses the signal and a single-stream transformer trained under a
flow-matching objective generates in the learned latent
space~\cite{rombach2022ldm,peebles2023dit,esser2024sd3,flux2024}. We present CONFLUX, a
natively 3D instance of this design for chest CT, conditioned on structured
radiological metadata (abnormality findings, sex, age, and reconstruction kernel)
through adaptive layer normalization. We show it is competitive
with a strong volumetric baseline on distribution-level quality while exposing
direct control over clinical attributes.

Flow matching trains the model so its outputs reproduce the data distribution for
each conditioning vector, an aggregate property that holds over many samples but is
not guaranteed for any single one. The objective rewards overall realism and never
checks whether an individual generated volume exhibits its requested findings, so a
sample can look realistic yet under-express or omit one. The gap is observable post hoc: a
classifier trained on real volumes reads requested attributes from generated
samples less reliably than from real ones, but the likelihood objective cannot
target it. Reinforcement learning (RL) optimizes it directly, maximizing an
explicit reward on the model's own samples. Group-relative policy optimization has
recently been adapted from language models to flow-matching image
generators~\cite{flowgrpo2025,dancegrpo2025}; we adapt it to 3D
structured-attribute medical synthesis as a post-training stage that improves
conditioning faithfulness, verified by an independent judge model.

\paragraph{Contributions.}
\begin{itemize}
\item \textbf{A controllable 3D latent rectified-flow model for chest CT.} A
  3D convolutional VAE and a single-stream rectified-flow transformer conditioned
  directly on structured radiological metadata through adaLN modulation, with
  synthesis quality competitive with a strong 3D CT diffusion baseline
  (Sec.~\ref{sec:vae}--\ref{sec:dit}, Sec.~\ref{sec:exp-quality}).
\item \textbf{RL post-training that improves conditioning faithfulness.} A
  group-relative policy optimization stage that fine-tunes the flow model against a
  faithfulness reward, improving how reliably requested findings appear in generated
  volumes. The gain is verified by an independent held-out judge; to our knowledge
  it is the first GRPO post-training of a 3D medical flow model (Sec.~\ref{sec:grpo},
  Sec.~\ref{sec:exp-faith}).
\item \textbf{A released model and synthetic dataset.} The trained model
  together with a ${\sim}200{,}000$-volume controllable, faithfulness-optimized
  synthetic chest-CT dataset whose conditioning metadata spans a wide variety of
  findings (Sec.~\ref{sec:exp-data-release}).
\end{itemize}

\section{Related Work}
\label{sec:related}

Latent diffusion compresses volumes with an autoencoder and learns a diffusion or
flow model over the latent representation~\cite{rombach2022ldm}. The image-generation
architecture has converged on the diffusion transformer~\cite{peebles2023dit},
trained under rectified-flow / flow-matching objectives~%
\cite{liu2022rectified,lipman2023flowmatching,esser2024sd3} with
adaptive-normalization conditioning, as
in recent open systems~\cite{flux2024}. In 3D medical imaging, conditional latent
diffusion has been applied to brain MRI~\cite{pinaya2022brainldm,vanpuyvelde2026braing3n}
and to whole-body and chest CT~\cite{guo2024maisi,wang2024meddiffusion3d,generatect}; MAISI, a 3D CT
latent diffusion model with mask- and metadata-based
conditioning~\cite{guo2024maisi}, is our reference competitor for synthesis
quality (Sec.~\ref{sec:exp-quality}).

Treating denoising as a multi-step decision process enables policy-gradient
fine-tuning of diffusion models against non-differentiable
rewards~\cite{black2023ddpo,fan2023dpok}. Group-relative policy optimization
(GRPO)~\cite{shao2024grpo}, which replaces a learned value function with a
group-relative advantage, was adapted to flow-matching generators by
Flow-GRPO~\cite{flowgrpo2025} via an ODE-to-SDE conversion, and to several visual
generation settings by DanceGRPO~\cite{dancegrpo2025}; subsequent work studies the
stability of the importance ratio in this regime~\cite{grpoguard2025}. Faithfulness,
the agreement between requested and realized attributes, is commonly measured by
scoring generated samples with a classifier trained on real
data~\cite{pinaya2022brainldm}.

\section{Method}
\label{sec:method}

\begin{figure}[t]
\centering
\includegraphics[width=\linewidth]{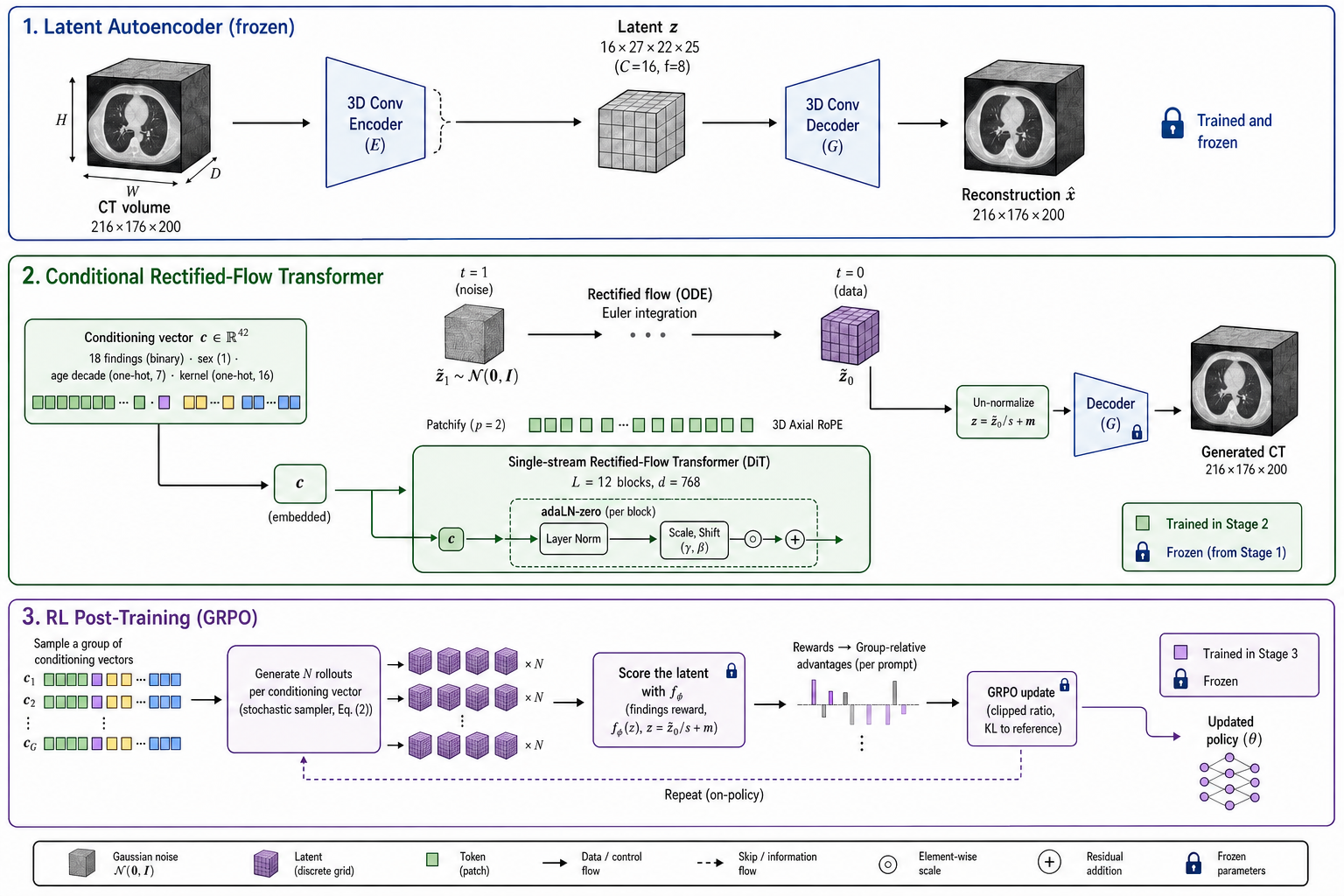}
\captionsetup{font=footnotesize}
\caption{CONFLUX architecture. \textbf{Stage~1:} a 3D convolutional VAE ($f{=}8$,
$C{=}16$; $216{\times}176{\times}200\!\to\!16{\times}27{\times}22{\times}25$) encodes
a CT volume to a latent representation and is frozen thereafter. \textbf{Stage~2:} a single-stream
rectified-flow transformer ($L{=}12$, $d{=}768$, patch size $p{=}2$) is trained from
scratch over the normalized latent space by flow matching; patchified tokens carry 3D axial
RoPE and are modulated block-wise by the structured metadata vector $\cond$
(Eq.~\eqref{eq:cond}) through adaLN-zero, and a sample is drawn by integrating the
Euler ODE from $t{=}1$ to $t{=}0$ and decoding with the frozen Stage-1 decoder.
\textbf{Stage~3:} GRPO post-training draws a group of rollouts per prompt with the
stochastic sampler (Eq.~\eqref{eq:sde}), scores each by the frozen findings-reward
classifier (Eq.~\eqref{eq:reward}), and turns the group-relative advantages into a
clipped, KL-anchored policy update against the frozen pre-RL reference.}
\label{fig:arch}
\end{figure}

CONFLUX is a three-stage latent diffusion model (Fig.~\ref{fig:arch}): a VAE compresses a CT volume to a low-resolution
latent representation (Sec.~\ref{sec:vae}); a single-stream rectified-flow
transformer generates in that latent space (Sec.~\ref{sec:dit}); and a third stage
post-trains this model to improve its faithfulness to the requested conditioning
(Sec.~\ref{sec:grpo}).

\subsection{Latent autoencoder}
\label{sec:vae}
Generating directly in voxel space is costly, so the first stage learns a compact
latent representation in which the flow model operates. A 3D convolutional variational autoencoder
encodes each preprocessed volume $\bm{x}\in\R^{1\times D\times H\times W}$ into a
diagonal-Gaussian latent $\latent\in\R^{C\times D/f\times H/f\times W/f}$ (mean and
per-element variance from an encoder $E$), downsampling by $f{=}8$ per spatial axis
to $C{=}16$ channels; a deeper decoder $G$ reconstructs $\hat{\bm{x}}=G(\latent)$.
Both use group-normalized residual blocks with a 3D self-attention block at the
lowest resolution, the encoder--decoder asymmetry following latent-diffusion
autoencoders~\cite{rombach2022ldm}. Training minimizes an $\ell_1$ reconstruction
loss, a Kullback--Leibler regularizer toward $\mathcal{N}(\bm 0,\bm I)$ (weight
$10^{-6}$), and a tri-planar LPIPS perceptual loss~\cite{zhang2018lpips} averaged
over slices along the three canonical planes (LPIPS is 2D). After training $E,G$
are frozen. Rectified-flow training assumes near-unit-variance targets, so the
flow model operates on the normalized latent $\tilde{\latent}=(\latent-m)\,s$ with
a single global scale $s$ and shift $m$ estimated from the training latents,
inverted as $\latent=\tilde{\latent}/s+m$ before decoding or scoring
(values in Table~\ref{tab:hparams}).

\subsection{Conditional rectified-flow transformer}
\label{sec:dit}
The second stage is a single-stream transformer that generates a latent embedding from a
Gaussian noise sample. Under the rectified-flow formulation, a data
latent $\tilde{\latent}_0$ and noise $\bm\epsilon\sim\mathcal{N}(\bm 0,\bm I)$ are
joined by the straight-line path
$\tilde{\latent}_t = (1-t)\,\tilde{\latent}_0 + t\,\bm\epsilon$ ($t{=}0$ data,
$t{=}1$ noise), travelled at the constant velocity $\bm\epsilon-\tilde{\latent}_0$.
The network $\vt(\tilde{\latent}_t,t,\cond)$ is trained to predict this velocity by
minimizing its expected squared error against the target
$\bm\epsilon-\tilde{\latent}_0$, and at sampling time the predicted velocity field
is integrated from noise back to a data latent. The latent volume is
patchified into tokens with 3D axial rotary position embeddings;
conditioning enters through adaptive layer-normalization (adaLN-zero). The
conditioning vector concatenates the available
structured metadata,
\begin{equation}
\cond = \big[\, \bm{c}_{\mathrm{find}} \in \{0,1\}^{18},\;
c_{\mathrm{sex}} \in \{0,1\},\;
\bm{c}_{\mathrm{age}} \in \Delta^{6},\;
\bm{c}_{\mathrm{ker}} \in \Delta^{15} \,\big] \in \R^{42},
\label{eq:cond}
\end{equation}
i.e.\ $18$ binary findings, sex, a one-hot age decade ($\Delta^{k}$ the
$k$-simplex vertices), and a one-hot reconstruction kernel. For classifier-free
guidance~\cite{ho2022cfg} $\cond$ is dropped to a null embedding with probability
$0.1$ during training; sampling integrates the probability-flow ODE
$\mathrm{d}\tilde{\latent}=\vt\,\mathrm{d}t$ from $t{=}1$ to $t{=}0$ on a
time-shifted Euler grid. Training uses precomputed latent moments, so $E$ is never
evaluated in stage two.

\subsection{Reinforcement-learning post-training}
\label{sec:grpo}
Flow matching trains the generator to match the data distribution, but supplies no
signal that a particular requested attribute is realized in a given sample. The
post-training stage adds that signal: it samples volumes from the model, scores how
well each matches its requested conditioning, and updates the model to increase that
agreement, adapting Flow-GRPO~\cite{flowgrpo2025} to this setting. Policy-gradient
updates require a probability for each sampling step, which the deterministic ODE
sampler does not provide, so we replace it with a stochastic sampler that preserves
the model's marginals while injecting noise at each step,
\begin{equation}
\tilde{\latent}_{t+\Delta t}
= \tilde{\latent}_t + \Delta t\,\bm{d}(\tilde{\latent}_t,t)
+ \sigma(t)\sqrt{\lvert\Delta t\rvert}\;\bm\xi,
\qquad
\bm{d} = \vt + \frac{\sigma(t)^2}{2t}\big(\tilde{\latent}_t + (1-t)\vt\big),
\label{eq:sde}
\end{equation}
($\Delta t<0$, action noise $\bm\xi\sim\mathcal{N}(\bm 0,\bm I)$). Each step is now
a Gaussian draw, giving a $T$-step rollout a tractable log-probability. Its per-step
term is averaged over the $d{\approx}2.4{\times}10^5$ latent dimensions rather than
summed: summation over so many dimensions inflates the ratio between the updated and
sampling policies and destabilizes training, a known failure mode of flow-model RL.
The reward measures how faithfully a generated volume realizes its requested
findings, read by a frozen classifier $f_\phi$,
\begin{equation}
r(\tilde{\latent}_0,\cond)
= -\!\!\sum_{g\in\{\mathrm{find},\mathrm{sex},\mathrm{age},\mathrm{ker}\}}\!\!
\omega_g\,\ell_g\big(f_\phi(\latent),\,\cond_g\big),
\qquad
\latent = \tilde{\latent}_0/s + m,
\label{eq:reward}
\end{equation}
the negative weighted cross-entropy between predicted and requested conditioning,
with the latent un-normalized first because $f_\phi$ is trained on raw autoencoder
latents. This classifier is a compact 3D convolutional network (group-normalized
convolutional blocks, global average pooling, and a linear head over the $18$
findings) trained on real latent representations; the independent judge used in evaluation
(Sec.~\ref{sec:exp-faith}) shares this architecture but scores decoded volumes at the voxel level. We
optimize the findings group only ($\omega_{\mathrm{find}}{=}1$, the others $0$);
the unweighted groups serve as a specificity check (Sec.~\ref{sec:experiments}). For each
step we draw $G$ conditioning vectors and $N$ rollouts per vector and form
group-relative advantages in the standard way, subtracting each prompt's mean
reward and dividing by the batch-wide standard deviation (more stable than a
per-group estimate at the small $N$ affordable here). The policy is updated with
the usual clipped PPO surrogate on the per-step importance ratio
$\policy/\policy^{\mathrm{old}}$, penalized by a Kullback--Leibler divergence to the
frozen pre-RL reference $\refpolicy$. With one on-policy epoch the clip is inactive
and this penalty is the sole brake on drift; the
configuration is given in Table~\ref{tab:hparams}.

\section{Experiments}
\label{sec:experiments}

We use CT-RATE~\cite{ctrate2024}, a chest-CT corpus with paired structured
metadata and $18$ NLP-extracted abnormality labels. Volumes are cropped to a lung
bounding box and resized to $216{\times}176{\times}200$ with intensities in
$\sim[-1,1]$. A quality-control pass removes scans failing thresholds on lung
fraction, left--right balance, and voxel spacing, and removes feet-first and
bone-kernel acquisitions; the autoencoder trains on the QC-passing set
(${\sim}40{,}800$ volumes) and the flow model on a one-scan-per-patient subset of
$18{,}417$, with $3{,}039$ patient-disjoint volumes held out for validation. Because the $18$ findings are NLP-extracted rather than
expert-annotated, faithfulness is measured against these predicted labels.

\subsection{Synthesis quality}
\label{sec:exp-quality}

\begin{figure}[t]
\centering
\includegraphics[width=\linewidth]{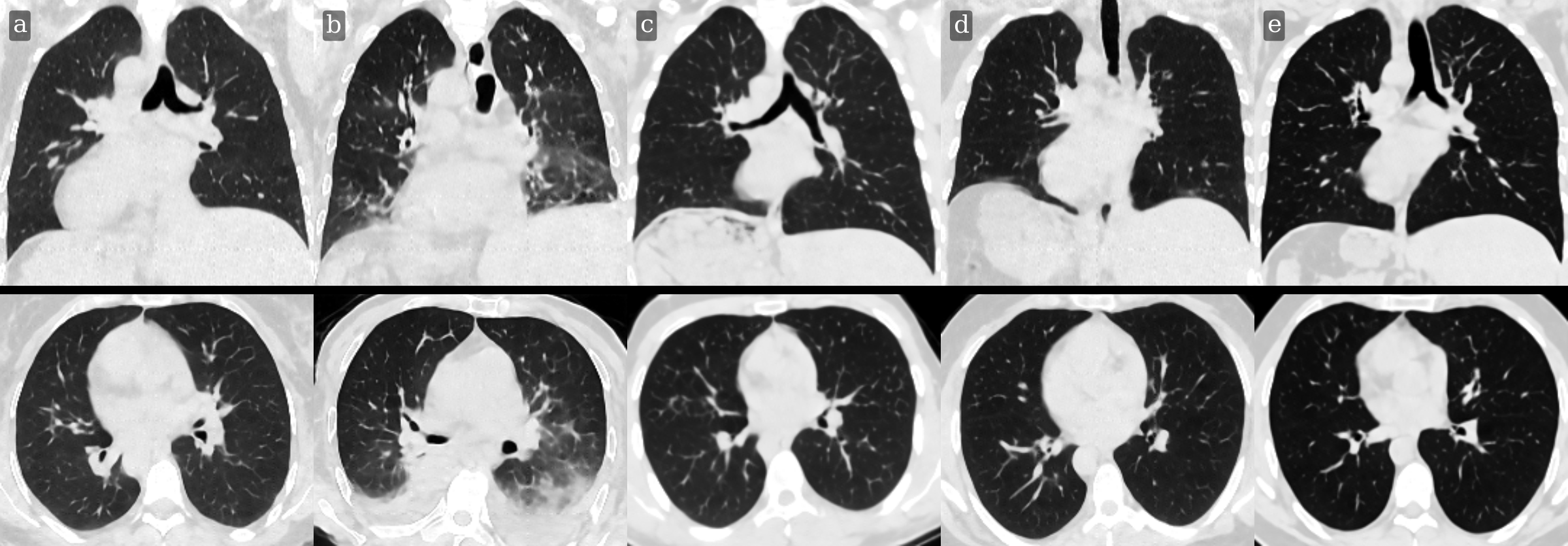}
\caption{CONFLUX samples. Generated volumes conditioned on real CT-RATE metadata;
coronal (top) and axial (bottom) mid-slices, lung window. \textbf{(a)}~F,
$40$--$50$: cardiomegaly, lung nodule, mosaic attenuation; \textbf{(b)}~M, 70+:
pleural effusion, opacity, consolidation; \textbf{(c)}~M, $20$--$30$ and
\textbf{(d)}~M, $30$--$40$: no abnormality; \textbf{(e)}~M, $30$--$40$: lung nodule,
bronchiectasis. Not all listed findings appear in the shown mid-slice. More samples
with per-volume conditioning are in Appendix~\ref{sec:appendix}.}
\label{fig:samples}
\end{figure}

We measure distribution-level quality with a tri-planar 2D-FID (Inception-v3 pool3
features on evenly spaced axial/coronal/sagittal slices, averaged), complemented by density
and coverage~\cite{naeem2020} on the same tri-planar features ($k{=}5$; fidelity
vs.\ manifold coverage) and diversity (mean pairwise MS-SSIM among generated
volumes) against the real floor. Each method is scored on $N{=}501$ generated volumes against the held-out
validation reference, with identical lung windowing and seed. Competitors are MAISI~\cite{guo2024maisi}, a 3D CT latent
diffusion model, and GenerateCT~\cite{generatect}, a text-conditional 3D chest-CT
generator; VAE reconstruction upper-bounds any latent model on our autoencoder.
Fig.~\ref{fig:samples} shows generated volumes across a range of conditioning
profiles, with matching coronal and axial views.

Our model leads both baselines (Table~\ref{tab:quality}): tri-planar FID
$32.3$ against $74.6$ (MAISI) and $145.4$ (GenerateCT), approaching
the $22.6$ VAE-reconstruction ceiling. Every FID and density/coverage gap to ours
is significant (unpaired bootstrap, $95\%$ CIs exclude zero); our samples are
$3.5\times$ denser and cover $6\times$ more of the real manifold than MAISI.
Diversity sits near the real floor for all methods ($0.51$ ours vs.\ $0.41$ real;
collapse would approach $1$). Competitors are resampled into our lung space,
inflating their cross-plane FID, so the resize-robust axial plane is the cleanest
comparison and still favors ours ($24.7$ vs.\ $55.6$ and $70.2$).

\begin{table}[tb]
\centering
\setlength{\tabcolsep}{3.2pt}
\caption{Synthesis quality vs.\ 3D CT baselines ($N{=}501$; $\pm95\%$ bootstrap
CI). Tri-planar and axial 2D-FID ($\downarrow$), density and
coverage~\cite{naeem2020} ($\uparrow$), and diversity (pairwise MS-SSIM, real floor
$0.41$). VAE reconstruction is the autoencoder upper bound; best generative result
in bold.}
\label{tab:quality}
\begin{tabular}{lccccc}
\toprule
Method & FID$_{\text{tri}}\downarrow$ & FID$_{\text{ax}}\downarrow$ & Density $\uparrow$ & Coverage $\uparrow$ & Diversity \\
\midrule
GenerateCT~\cite{generatect} & $145.4{\pm}2.3$ & $70.2{\pm}3.1$ & $0.008{\pm}.002$ & $0.009{\pm}.002$ & $0.486$ \\
MAISI~\cite{guo2024maisi} & $74.6{\pm}3.3$ & $55.6{\pm}2.9$ & $0.016{\pm}.003$ & $0.017{\pm}.003$ & $0.471$ \\
\textbf{CONFLUX (ours)} & $\mathbf{32.3{\pm}1.9}$ & $\mathbf{24.7{\pm}1.3}$ & $\mathbf{0.056{\pm}.006}$ & $\mathbf{0.105{\pm}.008}$ & $0.507$ \\
VAE recon (ceiling) & $22.6{\pm}1.8$ & $18.9{\pm}1.3$ & $0.161{\pm}.010$ & $0.466{\pm}.015$ & $0.453$ \\
\bottomrule
\end{tabular}
\end{table}

\subsection{Reward classifier}
\label{sec:exp-clf}

The GRPO reward is a frozen classifier that reads the $18$ findings from the latent
(Sec.~\ref{sec:grpo}), so its accuracy bounds the quality of the reward signal. On
the real validation split it reaches macro AUROC $0.793$ over the $18$ labels,
above every CT-CLIP variant reported on the same CT-RATE
benchmark~\cite{ctrate2024} (Table~\ref{tab:clf}), outperforming CT-CLIP zero-shot on $17/18$ findings. It does so while scoring the $8{\times}$-compressed latent rather
than full-resolution CT, indicating the tokenizer preserves the diagnostic signal.
The comparison is not fully controlled (we score a one-scan-per-patient subset from
the latent; the baselines, the full validation split at full resolution), but the
dataset, labels, findings, and metric are identical.

\begin{table}[tb]
\centering
\small
\setlength{\tabcolsep}{5pt}
\caption{Per-finding AUROC on CT-RATE (real validation): our latent classifier
vs.\ CT-CLIP variants and CT-Net~\cite{ctrate2024}. Top $8$ findings by our AUROC;
macro over all $18$ (full table in Table~\ref{tab:clf-full}). Best per row in bold.}
\label{tab:clf}
\begin{tabular}{lcccc}
\toprule
Finding & Ours & CT-Net & CT-CLIP$_{\text{zs}}$ & CT-CLIP$_{\text{ft}}$ \\
\midrule
Cardiomegaly & \textbf{0.92} & 0.79 & 0.86 & \textbf{0.92} \\
Pleural effusion & 0.91 & 0.77 & 0.90 & \textbf{0.93} \\
Arterial wall calcif. & \textbf{0.91} & 0.63 & 0.85 & 0.87 \\
Coronary artery calcif. & \textbf{0.91} & 0.63 & 0.85 & 0.86 \\
Interlobular septal thick. & \textbf{0.88} & 0.73 & 0.77 & 0.80 \\
Pericardial effusion & \textbf{0.83} & 0.65 & 0.77 & 0.78 \\
Mosaic attenuation & \textbf{0.81} & 0.74 & 0.77 & 0.78 \\
Bronchiectasis & \textbf{0.78} & 0.57 & 0.65 & 0.61 \\
\midrule
\textbf{Macro (all 18)} & \textbf{0.793} & 0.629 & 0.731 & 0.756 \\
\bottomrule
\end{tabular}
\end{table}

\subsection{Conditioning faithfulness}
\label{sec:exp-faith}

We measure faithfulness as agreement between requested and realized conditioning,
read by a classifier on the generated samples. The \emph{reward} classifier
operates in the latent space and is the model optimized by GRPO; because using it for
evaluation is circular, we judge instead with an \emph{independent image-space}
classifier trained on decoded real volumes and never used as the reward. We
generate from $200$ fixed validation prompts with $4$ samples each at $50$ ODE
steps and report macro average precision (AP) and macro AUROC over the $18$
findings. Post-training raises both under the independent judge
(Table~\ref{tab:faith}): AP $0.330{\to}0.344$ and AUROC $0.684{\to}0.699$. The
absolute gains are small, but so is the headroom: the judge reads \emph{real} CT at
only AP $0.360$, so the gain already recovers $47\%$ of the base-to-real gap. A paired-by-prompt bootstrap ($B{=}2000$), which
cancels prompt-difficulty noise, confirms both gains significant ($p{=}0.042$ and
$0.014$). The unweighted groups (sex, age, kernel) stay unchanged within noise, so the gain
is specific to the optimized findings.

\begin{table}[tb]
\centering
\caption{Conditioning faithfulness under the \emph{independent} image-space judge,
findings macro over $18$ labels ($n{=}800$ generated). The real-data ceiling is the
judge scored on real volumes; $\Delta$ and its $\pm1$ s.e.\ come from a
paired-by-prompt bootstrap, both gains significant. Headroom recovered is $\Delta$
as a fraction of the base-to-ceiling gap.}
\label{tab:faith}
\begin{tabular}{lcc}
\toprule
 & AP $\uparrow$ & AUROC $\uparrow$ \\
\midrule
Base (pre-RL) & $0.330$ & $0.684$ \\
\textbf{CONFLUX} & $\mathbf{0.344}$ & $\mathbf{0.699}$ \\
Real-data ceiling & $0.360$ & $0.746$ \\
\midrule
$\Delta$ (CONFLUX $-$ base) & $+0.014{\pm}0.008$ & $+0.015{\pm}0.006$ \\
Headroom recovered & $47\%$ & $24\%$ \\
\bottomrule
\end{tabular}
\end{table}

\subsection{Model and dataset release}
\label{sec:exp-data-release}

We release the trained model and, conditioning it on the real CT-RATE metadata
proportions, a ${\sim}200{,}000$-volume synthetic chest-CT dataset.
Each volume is paired with its conditioning vector, spanning a wide variety of
findings, for cohort augmentation and conditional study design at a scale
unavailable in real corpora. The dataset is available at
\url{https://huggingface.co/datasets/gevaertlab/conflux-chest-ct} and the model
checkpoints at \url{https://huggingface.co/gevaertlab/conflux}.

\section{Conclusion}
\label{sec:conclusion}

We presented CONFLUX, a controllable, natively 3D latent rectified-flow model
for chest CT, conditioned on structured radiological metadata, that leads
strong 3D baselines on quality while giving direct control over clinical
attributes. An RL post-training stage raises conditioning faithfulness, the first GRPO post-training of a 3D
medical flow model to our knowledge. We release the model and a
${\sim}200{,}000$-volume synthetic chest-CT dataset.

\bibliographystyle{splncs04}
\bibliography{refs}

\clearpage
\appendix
\section{Training Parameters}
\label{hparams}

\begin{table}[!ht]
\centering
\small
\caption{Training configuration for the three stages. Latent moments are cached;
the post-training SDE uses the FLUX time-shifted Euler grid; the reward classifier
and reference policy are frozen. Per-channel latent std is $1.27$--$2.48$.}
\label{tab:hparams}
\setlength{\tabcolsep}{6pt}
\begin{tabular}{lll}
\toprule
Stage & Hyperparameter & Value \\
\midrule
Autoencoder & downsampling factor $f$ & $8$ \\
 & latent channels $C$ & $16$ \\
\midrule
Flow transformer & hidden width $d$ & $768$ \\
 & depth $L$ & $12$ \\
 & attention heads & $12$ \\
 & patch size $p$ & $2$ \\
 & latent norm.\ scale $s$ & $0.5509$ \\
 & latent norm.\ shift $m$ & $0.1869$ \\
\midrule
GRPO post-training & prompts per step $G$ & $2$ \\
 & rollouts per prompt $N$ & $16$ \\
 & SDE steps $T$ & $10$ \\
 & action noise $\sigma(t)$ & $0.5\sqrt{t}$ \\
 & PPO clip $\epsilon$ & $0.2$ \\
 & KL weight $\beta_{\KL}$ & $0.03$ \\
 & inner epochs & $1$ \\
 & learning rate & $3{\times}10^{-6}$ \\
\bottomrule
\end{tabular}
\end{table}

\clearpage
\section{Per-finding reward-classifier accuracy}
\label{app:clf}

\begin{table}[!ht]
\centering
\small
\setlength{\tabcolsep}{5pt}
\caption{Per-finding AUROC on CT-RATE (real validation), all $18$ findings: our
latent classifier vs.\ CT-CLIP variants and CT-Net~\cite{ctrate2024}. Sorted by our
AUROC; best per row in bold.}
\label{tab:clf-full}
\begin{tabular}{lcccc}
\toprule
Finding & Ours & CT-Net & CT-CLIP$_{\text{zs}}$ & CT-CLIP$_{\text{ft}}$ \\
\midrule
Cardiomegaly & \textbf{0.92} & 0.79 & 0.86 & \textbf{0.92} \\
Pleural effusion & 0.91 & 0.77 & 0.90 & \textbf{0.93} \\
Arterial wall calcif. & \textbf{0.91} & 0.63 & 0.85 & 0.87 \\
Coronary artery calcif. & \textbf{0.91} & 0.63 & 0.85 & 0.86 \\
Interlobular septal thick. & \textbf{0.88} & 0.73 & 0.77 & 0.80 \\
Pericardial effusion & \textbf{0.83} & 0.65 & 0.77 & 0.78 \\
Mosaic attenuation & \textbf{0.81} & 0.74 & 0.77 & 0.78 \\
Bronchiectasis & \textbf{0.78} & 0.57 & 0.65 & 0.61 \\
Atelectasis & \textbf{0.76} & 0.61 & 0.68 & 0.69 \\
Consolidation & \textbf{0.76} & 0.66 & 0.72 & 0.75 \\
Medical material & \textbf{0.76} & 0.66 & 0.72 & 0.75 \\
Lung opacity & \textbf{0.76} & 0.60 & 0.63 & 0.69 \\
Hiatal hernia & \textbf{0.75} & 0.55 & 0.70 & 0.72 \\
Emphysema & 0.73 & 0.54 & 0.74 & \textbf{0.75} \\
Lymphadenopathy & \textbf{0.72} & 0.61 & 0.70 & 0.70 \\
Peribronchial thick. & \textbf{0.71} & 0.53 & 0.69 & \textbf{0.71} \\
Pulmonary fibrotic seq. & \textbf{0.70} & 0.52 & 0.57 & 0.65 \\
Lung nodule & \textbf{0.67} & 0.55 & 0.57 & 0.65 \\
\midrule
\textbf{Macro (all 18)} & \textbf{0.793} & 0.629 & 0.731 & 0.756 \\
\bottomrule
\end{tabular}
\end{table}

\clearpage
\section{Additional samples and conditioning}
\label{sec:appendix}

\begin{figure}[!ht]
\centering
\includegraphics[width=0.9\textwidth]{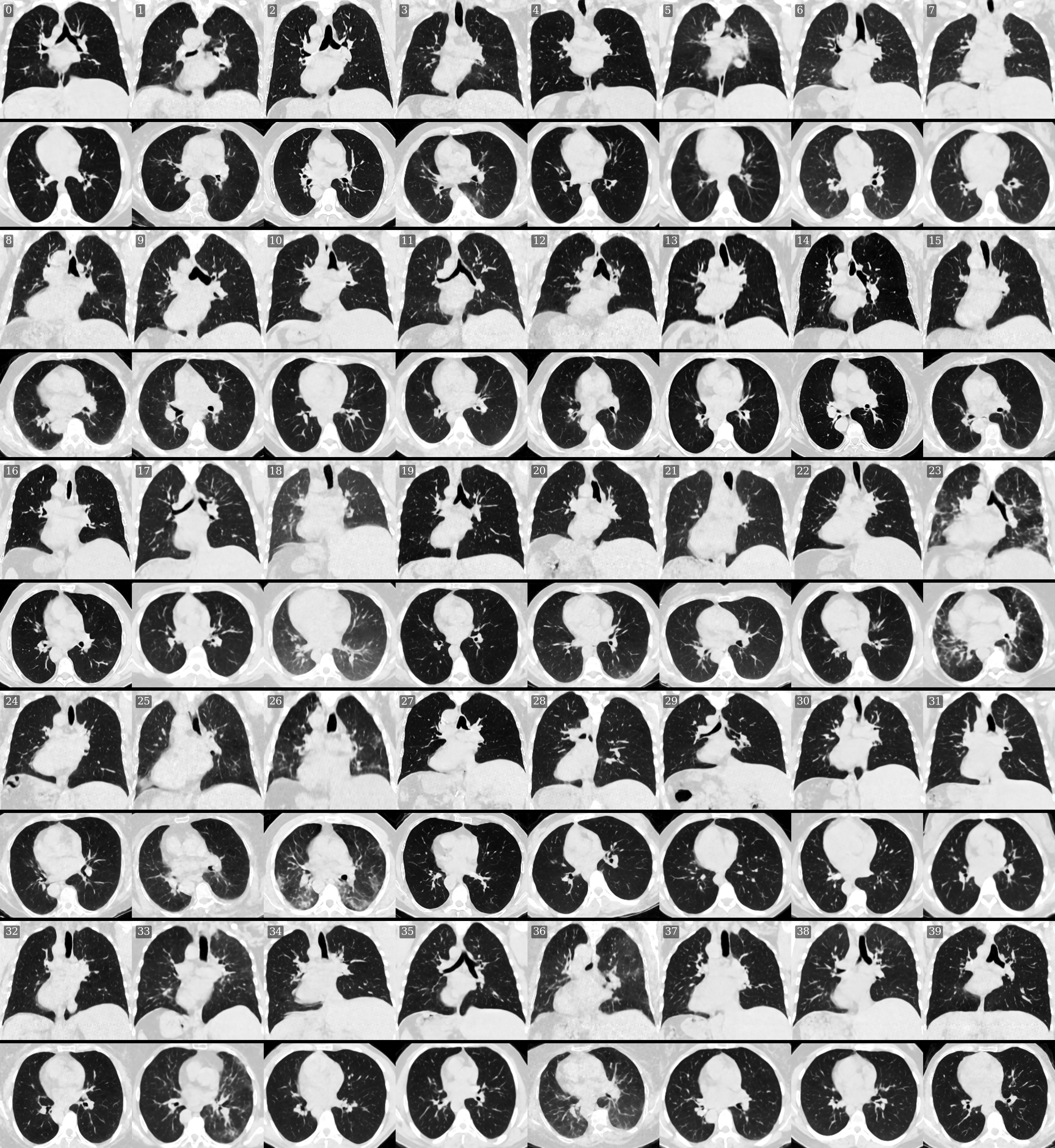}
\caption{CONFLUX samples. Forty synthetic chest CT volumes drawn at
random from the model, each as a coronal (top) and axial (bottom)
mid-slice in a lung window, numbered $0$--$39$; per-patient conditioning is given
in Table~\ref{tab:appendix-meta}. The range of body habitus, anatomy, and
abnormality reflects the diversity and realism of generated volumes; matching
coronal/axial views show cross-plane 3D coherence.}
\label{fig:appendix}
\end{figure}

\clearpage
\begin{longtable}{r l l l p{7cm}}
\caption{Conditioning metadata for each synthetic volume in Fig.~\ref{fig:appendix}.}\label{tab:appendix-meta}\\
\toprule
\# & Sex & Age & Kernel & Conditioned abnormalities \\ \midrule
\endfirsthead
\toprule
\# & Sex & Age & Kernel & Conditioned abnormalities \\ \midrule
\endhead
0 & F & 0--20 & Br40f & none \\
1 & F & 70+ & YA & Arterial wall calcification, Cardiomegaly, Coronary artery wall calcification, Lymphadenopathy, Atelectasis, Lung nodule, Lung opacity \\
2 & M & 60--70 & Bl56f & Arterial wall calcification, Coronary artery wall calcification, Hiatal hernia, Pulmonary fibrotic sequela, Bronchiectasis \\
3 & M & 30--40 & YA & Lung opacity, Consolidation, Interlobular septal thickening \\
4 & M & 30--40 & EA & Hiatal hernia, Atelectasis \\
5 & F & 20--30 & Br40f & none \\
6 & M & 30--40 & Br40f & Lymphadenopathy, Peribronchial thickening, Consolidation, Bronchiectasis \\
7 & F & 20--30 & A & Atelectasis, Lung nodule \\
8 & M & 60--70 & YA & Arterial wall calcification, Lung opacity \\
9 & F & 40--50 & YA & Arterial wall calcification, Hiatal hernia, Lymphadenopathy, Atelectasis, Lung nodule, Pulmonary fibrotic sequela, Consolidation \\
10 & F & 30--40 & EA & Lymphadenopathy, Lung nodule \\
11 & M & 40--50 & YA & Lymphadenopathy, Lung nodule, Lung opacity \\
12 & M & 50--60 & YB & Lymphadenopathy, Lung nodule \\
13 & F & 50--60 & L & Coronary artery wall calcification, Lung nodule, Lung opacity, Pulmonary fibrotic sequela \\
14 & F & 70+ & Bl56f & Arterial wall calcification, Coronary artery wall calcification, Emphysema, Atelectasis, Lung nodule, Pulmonary fibrotic sequela, Mosaic attenuation pattern \\
15 & F & 70+ & YA & Arterial wall calcification, Coronary artery wall calcification, Atelectasis \\
16 & M & 50--60 & Bl57d & Lymphadenopathy, Lung opacity \\
17 & M & 20--30 & Br36d & none \\
18 & M & 20--30 & EA & none \\
19 & F & 40--50 & YA & Arterial wall calcification, Lymphadenopathy, Emphysema, Lung nodule, Lung opacity, Pulmonary fibrotic sequela \\
20 & M & 50--60 & YA & Lung opacity \\
21 & F & 40--50 & YA & none \\
22 & M & 40--50 & EA & Lymphadenopathy, Lung nodule \\
23 & F & 50--60 & Br40f & Cardiomegaly, Lymphadenopathy, Consolidation \\
24 & F & 60--70 & YA & Arterial wall calcification, Lymphadenopathy, Lung nodule, Pulmonary fibrotic sequela, Mosaic attenuation pattern \\
25 & F & 70+ & YA & Coronary artery wall calcification \\
26 & M & 50--60 & L & Lung opacity, Consolidation \\
27 & M & 70+ & Bl56f & Arterial wall calcification, Coronary artery wall calcification, Atelectasis, Lung opacity, Pulmonary fibrotic sequela \\
28 & F & 20--30 & YA & none \\
29 & M & 30--40 & EA & Lung nodule, Pulmonary fibrotic sequela \\
30 & F & 50--60 & Br40f & Arterial wall calcification, Coronary artery wall calcification, Hiatal hernia, Lung nodule \\
31 & F & 20--30 & B & none \\
32 & F & 50--60 & YA & Lymphadenopathy, Atelectasis, Lung opacity, Pulmonary fibrotic sequela \\
33 & F & 60--70 & Br36d & Lung opacity, Pulmonary fibrotic sequela, Peribronchial thickening, Consolidation, Bronchiectasis, Interlobular septal thickening \\
34 & M & 20--30 & EA & none \\
35 & M & 30--40 & B & Lung nodule \\
36 & F & 70+ & YA & Arterial wall calcification, Coronary artery wall calcification, Emphysema, Consolidation \\
37 & F & 40--50 & Br40f & none \\
38 & M & 20--30 & Br40f & none \\
39 & F & 30--40 & YA & none \\
\bottomrule
\end{longtable}

\clearpage
\begin{figure}[!ht]
\centering
\includegraphics[width=0.9\textwidth]{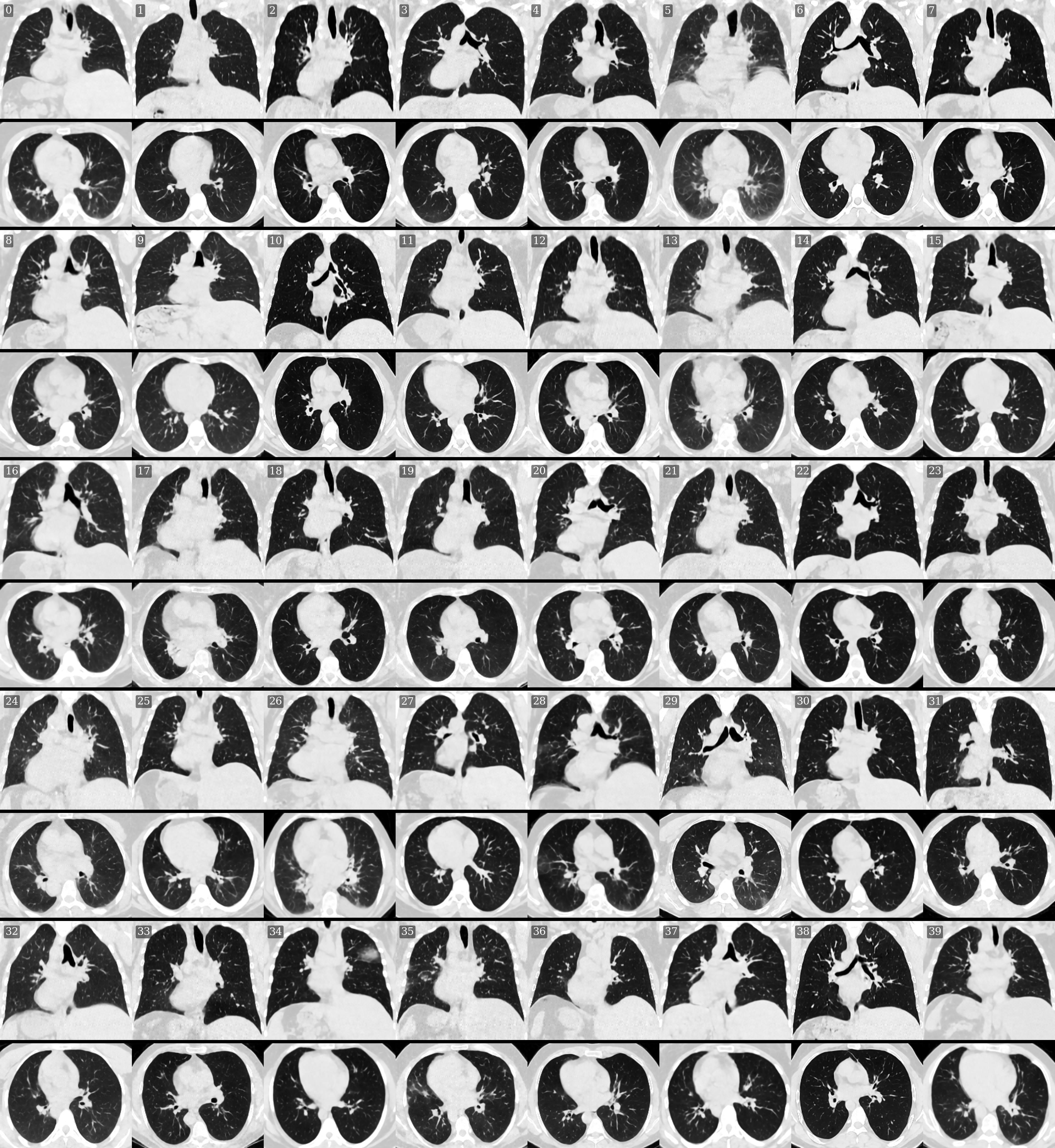}
\caption{CONFLUX samples (continued). A further $40$ synthetic chest CT
volumes drawn at random, each a coronal (top) and axial (bottom) mid-slice in a lung
window, numbered $0$--$39$; per-patient conditioning is given in
Table~\ref{tab:appendix-meta2}.}
\label{fig:appendix2}
\end{figure}

\clearpage
\begin{longtable}{r l l l p{7cm}}
\caption{Conditioning metadata for each synthetic volume in Fig.~\ref{fig:appendix2}.}\label{tab:appendix-meta2}\\
\toprule
\# & Sex & Age & Kernel & Conditioned abnormalities \\ \midrule
\endfirsthead
\toprule
\# & Sex & Age & Kernel & Conditioned abnormalities \\ \midrule
\endhead
0 & M & 40--50 & B & Medical material, Atelectasis, Lung nodule, Lung opacity \\
1 & F & 20--30 & YB & Lung nodule \\
2 & M & 50--60 & L & Arterial wall calcification, Atelectasis \\
3 & M & 20--30 & YA & Lung opacity, Interlobular septal thickening \\
4 & M & 50--60 & EA & Hiatal hernia, Atelectasis, Lung nodule, Pulmonary fibrotic sequela \\
5 & M & 50--60 & EA & Arterial wall calcification, Lung opacity \\
6 & F & 20--30 & Bl56f & Arterial wall calcification, Cardiomegaly, Pericardial effusion, Coronary artery wall calcification, Hiatal hernia, Lymphadenopathy, Lung nodule \\
7 & M & 30--40 & Bl57d & Emphysema, Atelectasis \\
8 & F & 50--60 & Br40f & none \\
9 & F & 30--40 & Br40f & none \\
10 & M & 40--50 & Bl56f & Hiatal hernia, Lymphadenopathy, Atelectasis, Lung nodule, Bronchiectasis \\
11 & F & 30--40 & Bl57d & none \\
12 & M & 50--60 & L & Lymphadenopathy, Lung opacity \\
13 & M & 50--60 & EA & Lung nodule \\
14 & F & 30--40 & Br60f & Medical material, Emphysema, Pulmonary fibrotic sequela \\
15 & F & 30--40 & Br40f & none \\
16 & M & 40--50 & B & Emphysema, Lung opacity, Consolidation \\
17 & F & 70+ & YA & Emphysema, Atelectasis, Lung nodule, Peribronchial thickening \\
18 & F & 60--70 & YA & Arterial wall calcification, Hiatal hernia, Atelectasis, Lung opacity \\
19 & F & 40--50 & YA & Emphysema \\
20 & F & 70+ & Br40f & Arterial wall calcification, Coronary artery wall calcification, Lymphadenopathy, Lung nodule, Pulmonary fibrotic sequela \\
21 & M & 20--30 & YA & Lung opacity, Peribronchial thickening, Bronchiectasis \\
22 & F & 40--50 & Br40f & Coronary artery wall calcification, Lung nodule \\
23 & M & 50--60 & EA & Medical material, Coronary artery wall calcification, Lymphadenopathy \\
24 & F & 50--60 & YA & Arterial wall calcification, Lymphadenopathy, Emphysema, Atelectasis \\
25 & M & 20--30 & Br40f & Lymphadenopathy, Lung nodule, Lung opacity \\
26 & M & 60--70 & B & Arterial wall calcification, Cardiomegaly, Pericardial effusion, Lymphadenopathy, Emphysema, Atelectasis, Lung opacity, Pulmonary fibrotic sequela, Pleural effusion, Peribronchial thickening, Bronchiectasis \\
27 & F & 30--40 & Br40f & none \\
28 & M & 60--70 & B & Arterial wall calcification, Coronary artery wall calcification, Hiatal hernia, Lung opacity \\
29 & M & 30--40 & Bl56f & Lymphadenopathy, Lung opacity, Bronchiectasis \\
30 & M & 40--50 & Br40f & Medical material, Emphysema, Lung nodule, Pulmonary fibrotic sequela \\
31 & M & 20--30 & YB & none \\
32 & M & 20--30 & YA & Lung nodule \\
33 & M & 50--60 & YA & Hiatal hernia, Lung nodule, Pulmonary fibrotic sequela, Mosaic attenuation pattern, Peribronchial thickening \\
34 & M & 30--40 & A & Lung opacity, Interlobular septal thickening \\
35 & M & 40--50 & YA & Hiatal hernia, Lung opacity \\
36 & M & 30--40 & Br60f & Bronchiectasis \\
37 & F & 30--40 & Br40f & Medical material, Pericardial effusion, Atelectasis, Lung nodule, Pulmonary fibrotic sequela \\
38 & M & 30--40 & Bl56f & none \\
39 & M & 50--60 & B & Lung opacity \\
\bottomrule
\end{longtable}

\clearpage
\begin{figure}[!ht]
\centering
\includegraphics[width=0.9\textwidth]{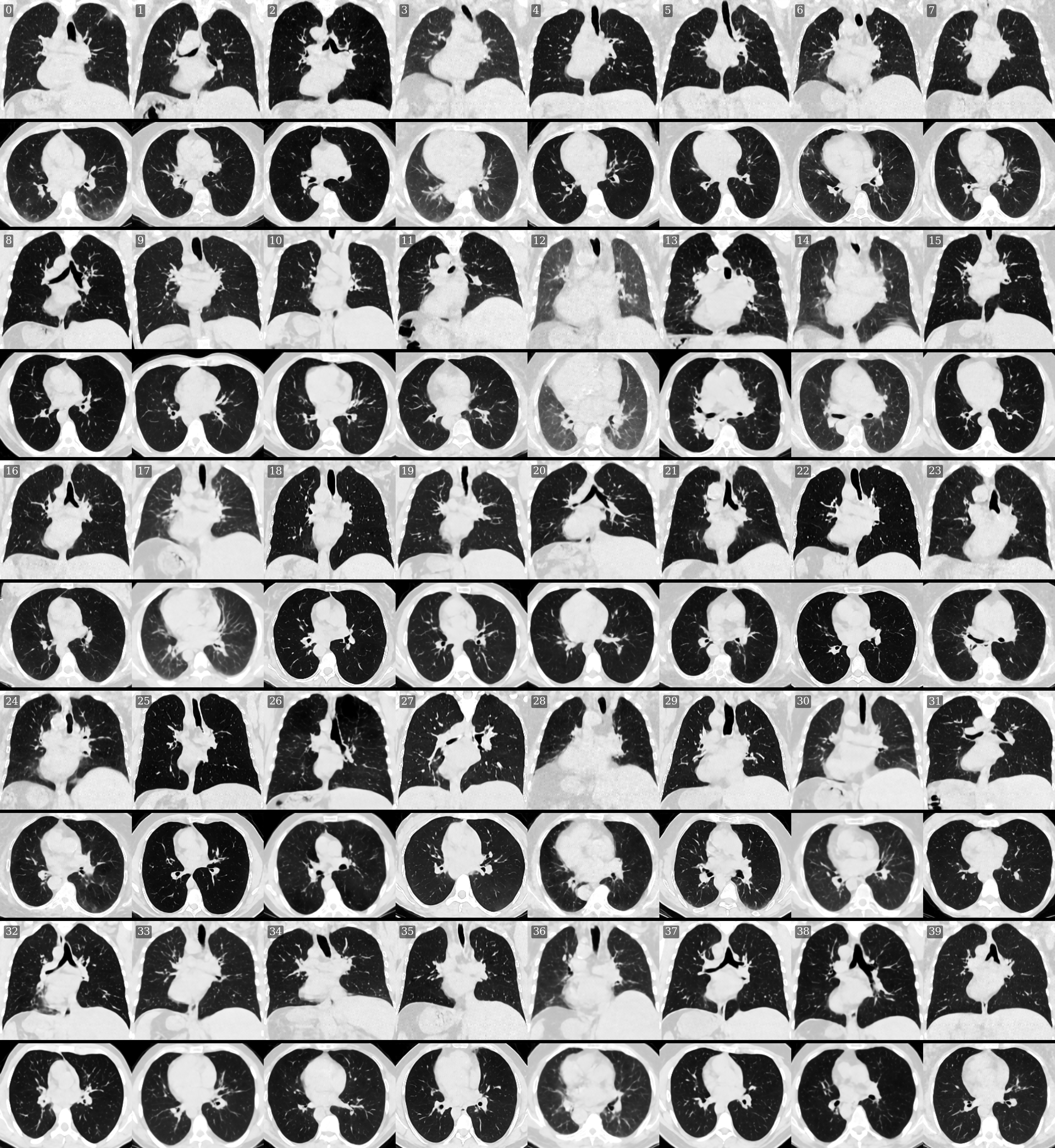}
\caption{CONFLUX samples (continued). A further $40$ synthetic chest CT
volumes drawn at random, each a coronal (top) and axial (bottom) mid-slice in a lung
window, numbered $0$--$39$; per-patient conditioning is given in
Table~\ref{tab:appendix-meta3}.}
\label{fig:appendix3}
\end{figure}

\clearpage
\begin{longtable}{r l l l p{7cm}}
\caption{Conditioning metadata for each synthetic volume in Fig.~\ref{fig:appendix3}.}\label{tab:appendix-meta3}\\
\toprule
\# & Sex & Age & Kernel & Conditioned abnormalities \\ \midrule
\endfirsthead
\toprule
\# & Sex & Age & Kernel & Conditioned abnormalities \\ \midrule
\endhead
0 & M & 30--40 & YA & Lung opacity \\
1 & F & 70+ & YA & Arterial wall calcification, Coronary artery wall calcification, Atelectasis, Lung nodule, Pulmonary fibrotic sequela \\
2 & M & 60--70 & YA & Arterial wall calcification, Coronary artery wall calcification, Emphysema, Pulmonary fibrotic sequela \\
3 & F & 50--60 & YA & Lung nodule, Lung opacity, Pulmonary fibrotic sequela \\
4 & F & 40--50 & YA & Medical material, Lung nodule \\
5 & F & 20--30 & YA & Pericardial effusion, Pleural effusion, Peribronchial thickening, Consolidation \\
6 & F & 70+ & Bl56f & Arterial wall calcification, Coronary artery wall calcification, Hiatal hernia, Lymphadenopathy, Lung nodule, Lung opacity, Pulmonary fibrotic sequela \\
7 & M & 50--60 & YA & Arterial wall calcification, Coronary artery wall calcification, Lymphadenopathy, Lung nodule, Lung opacity \\
8 & M & 40--50 & EA & Lung nodule, Pulmonary fibrotic sequela \\
9 & M & 30--40 & EA & Emphysema, Lung nodule, Pulmonary fibrotic sequela \\
10 & M & 40--50 & Br40f & Hiatal hernia, Atelectasis \\
11 & F & 60--70 & YA & Lymphadenopathy \\
12 & M & 60--70 & YA & Medical material, Arterial wall calcification, Cardiomegaly, Coronary artery wall calcification, Lymphadenopathy, Emphysema, Lung nodule, Pleural effusion, Mosaic attenuation pattern \\
13 & M & 70+ & Br40f & Arterial wall calcification, Cardiomegaly, Coronary artery wall calcification, Lymphadenopathy, Emphysema, Lung opacity, Pleural effusion, Peribronchial thickening, Consolidation \\
14 & F & 60--70 & Br60f & Cardiomegaly, Lymphadenopathy, Atelectasis \\
15 & M & 30--40 & EA & Lung nodule \\
16 & F & 40--50 & YA & Medical material, Atelectasis, Lung opacity, Consolidation \\
17 & M & 40--50 & B & Coronary artery wall calcification, Pulmonary fibrotic sequela \\
18 & F & 60--70 & Bl56f & Atelectasis, Lung opacity \\
19 & M & 30--40 & Br40f & Lymphadenopathy, Emphysema \\
20 & M & 20--30 & other & none \\
21 & F & 70+ & Br60f & Arterial wall calcification, Coronary artery wall calcification, Hiatal hernia, Lymphadenopathy, Emphysema, Lung nodule, Lung opacity, Consolidation \\
22 & F & 50--60 & Bl56f & Lung nodule \\
23 & F & 70+ & YA & Coronary artery wall calcification, Hiatal hernia, Atelectasis, Pleural effusion \\
24 & F & 70+ & Br60f & Medical material, Arterial wall calcification, Coronary artery wall calcification, Lung nodule, Lung opacity \\
25 & M & 40--50 & Bl56f & Lung nodule, Lung opacity \\
26 & M & 50--60 & B & Arterial wall calcification, Coronary artery wall calcification, Lymphadenopathy, Emphysema, Lung nodule, Pulmonary fibrotic sequela \\
27 & F & 30--40 & Bl56f & Lymphadenopathy, Lung nodule, Pulmonary fibrotic sequela \\
28 & M & 70+ & YA & Arterial wall calcification, Coronary artery wall calcification, Hiatal hernia, Lymphadenopathy, Lung nodule, Lung opacity, Pulmonary fibrotic sequela \\
29 & M & 50--60 & Bl56f & Medical material, Arterial wall calcification, Coronary artery wall calcification, Hiatal hernia, Emphysema, Lung nodule, Pulmonary fibrotic sequela, Bronchiectasis \\
30 & M & 50--60 & B & Medical material, Consolidation, Bronchiectasis \\
31 & F & 40--50 & YA & Lung nodule, Lung opacity \\
32 & M & 20--30 & Br40f & Lung opacity, Consolidation \\
33 & M & 30--40 & B & Lymphadenopathy, Consolidation \\
34 & M & 50--60 & EA & Lung nodule \\
35 & M & 30--40 & Bl56f & Lymphadenopathy, Emphysema, Atelectasis, Lung nodule \\
36 & M & 60--70 & B & Arterial wall calcification, Coronary artery wall calcification, Atelectasis, Lung nodule, Mosaic attenuation pattern \\
37 & M & 30--40 & Br40f & none \\
38 & M & 60--70 & B & Lung nodule, Pulmonary fibrotic sequela \\
39 & F & 40--50 & YA & Emphysema, Lung nodule, Lung opacity, Pulmonary fibrotic sequela \\
\bottomrule
\end{longtable}

\end{document}